# Neural Taylor Approximations: Convergence and Exploration in Rectifier Networks

**David Balduzzi** [1] **Brian McWilliams** [2] **Tony Butler-Yeoman** [1]

## Abstract

Modern convolutional networks, incorporating rectifiers and max-pooling, are neither smooth nor convex; standard guarantees therefore do not apply. Nevertheless, methods from convex optimization such as gradient descent and Adam are widely used as building blocks for deep learning algorithms. This paper provides the first convergence guarantee applicable to modern convnets, which furthermore matches a lower bound for *convex* nonsmooth functions. The key technical tool is the *neural Taylor approximation* – a straightforward application of Taylor expansions to neural networks – and the associated Taylor loss. Experiments on a range of optimizers, layers, and tasks provide evidence that the analysis accurately captures the dynamics of neural optimization. The second half of the paper applies the Taylor approximation to isolate the main difficulty in training rectifier nets – that gradients are *shattered* – and investigates the hypothesis that, by exploring the space of *activation configurations* more thoroughly, adaptive optimizers such as RMSProp and Adam are able to converge to better solutions.

## 1. Introduction

Deep learning has achieved impressive performance on a range of tasks (LeCun et al., 2015). The workhorse underlying deep learning is gradient descent or backprop. Gradient descent has convergence guarantees in settings that are smooth, convex or both. However, ***modern convnets are neither smooth nor convex***. Every winner of the ImageNet classification challenge since 2012 has used rectifiers which are not smooth (Krizhevsky et al., 2012; Zeiler & Fergus, 2014; Simonyan & Zisserman, 2015; Szegedy et al., 2015; He et al., 2015). Even in convex settings, convergence for nonsmooth functions is *lower-bounded* by $1/\sqrt{N}$ (Bubeck, 2015).

[1]Victoria University of Wellington, New Zealand [2]Disney Research, Zürich, Switzerland. Correspondence to: David Balduzzi <dbalduzzi@gmail.com>.

 

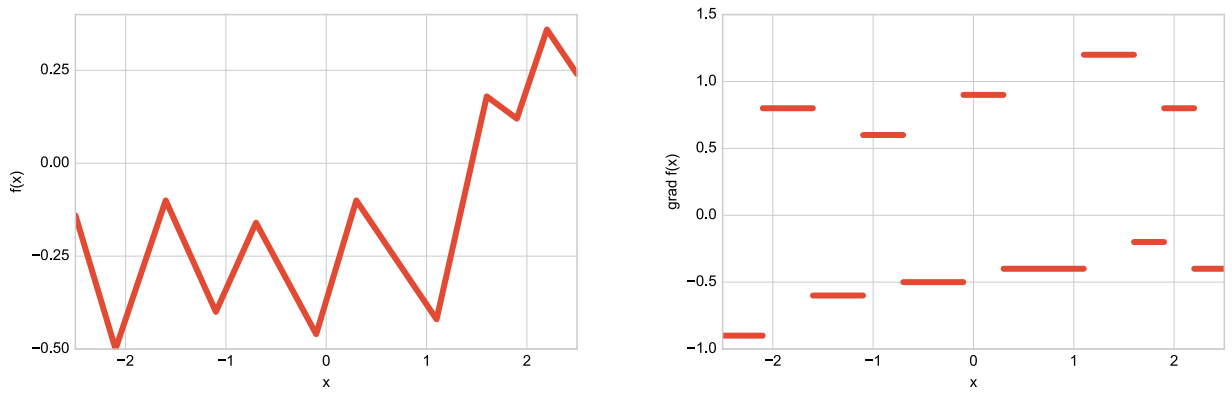

Fig. 1: **Shattered gradients in a PL-function.**

The paper's main contribution is the first convergence result for modern convnets, Theorem 2. The idea is simple: backprop constructs linear snapshots (gradients) of a neural net's landscape; section 2 introduces neural Taylor approximations which are used to construct Taylor losses as *convex snapshots* closely related to backprop. We then use the online convex optimization framework (Zinkevich, 2003) to show $1/\sqrt{N}$ convergence to the Taylor optimum, matching the lower bound in (Bubeck, 2015). Section 2.4 investigates the Taylor optimum and regret terms empirically. We observe that convergence to the Taylor optimum occurs at $1/\sqrt{N}$ in practice. The theorem applies to any neural net with a loss convex in the output of the net (for example, the cross-entropy loss is convex in the output but not the parameters of a neural net).

The nonsmoothness of rectifier nets is perhaps underappreciated (Balduzzi et al., 2017). Fig. 1 shows a piecewise-linear (PL) function and its gradient. The gradient is discontinuous or *shattered*. Shattering is problematic for accelerated and Hessian-based methods which speed up convergence by exploiting the relationship between gradients at nearby points (Sutskever et al., 2013). The success of these methods on *rectifier* networks, where the number of kinks grows exponentially with depth (Pascanu et al., 2014; Telgarsky, 2016), requires explanation since gradients at nearby points can be very different (Balduzzi et al., 2017).

Section 3 addresses the success of adaptive optimizers in rectifier nets.[1] **Adaptive optimizers** normalize gradients by their root-mean-square; e.g. AdaGrad, RMSProp, Adam and RadaGrad (Duchi et al., 2011; Hinton et al., 2012; Kingma & Ba, 2015; Krummenacher et al., 2016). Dauphin et al. (2015) argue that RMSProp approximates the equilibriation matrix $\sqrt{\text{diag}(H^2)}$ which approximates the absolute Hessian $|H|$ (Dauphin et al., 2014). However, the argument is at best part of the story when gradients are shattered. In fact, curvature-based explanations for RMS-normalization schemes do not tell the whole story even in smooth convex settings: Krummenacher et al. (2016) and Duchi et al. (2013) show that diagonal normalization schemes show no theoretical improvement over vanilla SGD when the coordinates are not axis-aligned or extremely sparse respectively.

The only way an optimizer can estimate gradients of a shattered function is to compute them directly. Effective optimizers must therefore *explore* the space of smooth regions – the bound in theorem 2 is only as good as the optimum over the Taylor losses encountered during backprop. Observations 1 and 2 relate smooth regions in rectifier nets and the Taylor losses to *configurations of active neurons*. We hypothesize that root-mean-square normalization increases exploration through the set of smooth regions in a rectifier net's landscape. Experiments in section 3.3 provide partial support for the hypothesis.

### 1.1. Comparison with related work

Researchers have applied convex techniques to neural networks. Bengio et al. (2006) show that choosing the number of hidden units converts neural optimization into a convex problem, see also Bach (2014). A convex multi-layer architectures are developed in Aslan et al. (2014); Zhang et al. (2016). However, these approaches have not achieved the practical success of convnets. In this work, we analyze convnets *as they are* rather than proposing a more tractable, but potentially less useful, model. A Taylor decomposition for neural networks was proposed in Montavon et al. (2015). They treat inputs as variable instead of weights and study interpretability instead of convergence. Taylor approximations to neural nets have also been used in Schraudolph (2002); Martens et al. (2012) to construct the generalized Gauss-Newton matrix as an alternative to the Hessian.

Our results are closely related to Balduzzi (2016), which uses game-theoretic techniques to prove convergence in rectifier nets. The approach taken here is more direct and holds in greater generality.

[1] For simplicity, we restrict to fully connected rectifier (ReLU) nets. The results also apply to convolutions, max-pooling, dropout, dropconnect, maxout, PReLUs and CReLUs (Srivastava et al., 2014; Wan et al., 2013; Goodfellow et al., 2013; He et al., 2015; Shang et al., 2016).

## 2. Convergence of Neural Networks

Before presenting the main result, we highlight some issues that arise when studying convergence in rectifier nets. Many optimization methods have guarantees that hold in convex or smooth settings. However, **none of the guarantees extend to rectifier nets**. For example, the literature provides no rigorous account of when or why Adam or Adagrad converges faster on rectifier nets than vanilla gradient descent. Instead, we currently have only intuition, empirics and an analogy with convex or smooth settings.

Gradient-based optimization on neural nets can converge on local optima that are substantially worse than the global optimum. Fortunately, "bad" local optima are rare in practice. A partial explanation for the prevalence of "good enough" local optima is Choromanska et al. (2015). Nevertheless, it is important to acknowledge that neural nets can and do converge to bad local optima. It is therefore impossible to prove (non-stochastic) bounds relative to the global optimum. Such a result may be provable under further assumptions. However, since the result would contradict empirical evidence, the assumptions would necessarily be unrealistic.

### 2.1. What kind of guarantee is possible?

The landscape of a rectifier net decomposes into smooth regions separated by kinks (Pascanu et al., 2014; Telgarsky, 2016), see figure 2. Gradients on different sides of a kink are unrelated since the derivative is discontinuous. Gradient-based optimizers cannot "peer around" the kinks in rectifier nets.

Gradient-based optimization on rectifier nets thus **decomposes** into two components. The first is steepest descent in a smooth region; the second moves between smooth regions. The first component is vanilla optimization whereas the second involves an element of **exploration**: what the optimizer encounters when it crosses a kink cannot be predicted.

The convergence guarantee in theorem 2 takes both components of the factorization into account in different ways. It is formulated in the adversarial setting of online convex optimization. Intuitively, the adversary is the nonsmooth geometry of the landscape, which generates what may-as-well-be a new loss whenever backprop enters a different smooth region.

Backprop searches a vast nonconvex landscape with a *linear flashlight* (the Taylor losses are a more sharply focused convex flashlight, see A4). The *adversary* is the landscape: from backprop's perspective its geometry – especially across kinks – is an unpredictable external force.

The Taylor losses are a series of convex problems that back-prop *de facto* optimizes – the gradients of the actual and Taylor losses are identical. The Taylor optimum improves when, stepping over a kink, backprop shines its light on a new (better) region of the landscape (fig. 2). Regret quantifies the gap between the Taylor optimal loss and the losses incurred during training.

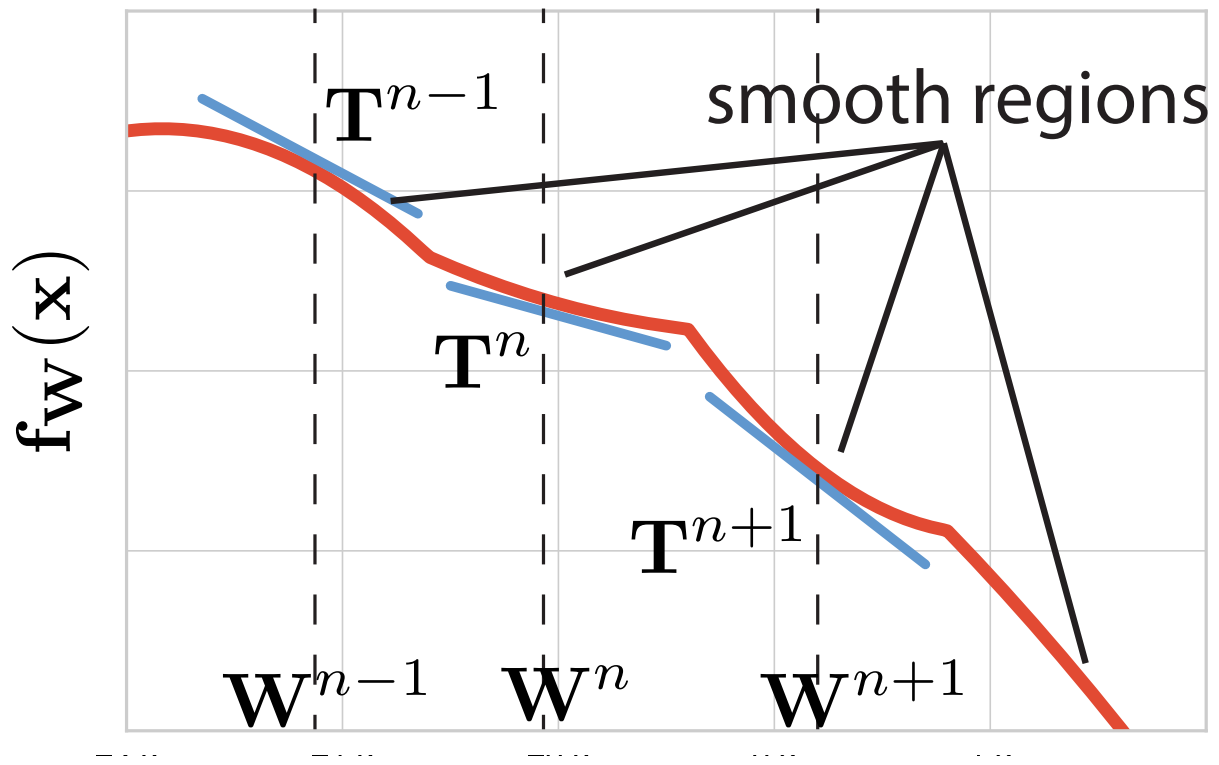

Fig. 2: **Neural Taylor approximation.**

### 2.2. Online Convex Optimization

In online convex optimization (Zinkevich, 2003), a learner is given convex loss functions $\ell^1, \dots \ell^N$. On the $n^{\text{th}}$ round, the learner predicts $\mathbf{W}^n$ *prior* to observing $\ell^n$, and then incurs loss $\ell^n(\mathbf{W}^n)$. Since the losses are not known in advance, the performance of the learner is evaluated *post hoc* via the **regret**, the difference between the incurred losses and the optimal loss in hindsight:

$$\texttt{Regret}(N) := \sum_{n=1}^{N} \Big[ \underbrace{\ell^n(\mathbf{W}^n)}_{\text{losses incurred}} - \underbrace{\ell^n(\mathbf{V}^*)}_{\text{optimal-in-hindsight}} \Big]$$

where $\mathbf{V}^* := \text{argmin}_{\mathbf{V}\in\mathcal{H}} \left[\sum_{n=1}^{N} \ell^n(\mathbf{V})\right]$. An algorithm has **no-regret** if $\lim_{N\to\infty} \texttt{Regret}(N)/N = 0$ for any sequence of convex losses with bounded gradients. For example, Kingma & Ba (2015) prove:

**Theorem 1** (Adam has no-regret)**.** *Suppose the convex losses $\ell^n$ have bounded gradients $\|\nabla_{\mathbf{W}} \ell^n(\mathbf{W})\|_2 \leq G$ and $\|\nabla_{\mathbf{W}} \ell^n(\mathbf{W})\|_\infty \leq G$ for all $\mathbf{W} \in \mathcal{H}$ and suppose that the weights chosen by the algorithm satisfy $\|\mathbf{W}^m - \mathbf{W}^n\|_2 \leq D$ and $\|\mathbf{W}^m - \mathbf{W}^n\|_\infty \leq D$ for all $m, n \in \{1, \dots, N\}$. Then Adam satisfies*

$$\texttt{Regret}(N)/N \leq O\left(1/\sqrt{N}\right) \quad \textit{for all } N \geq 1. \tag{1}$$

The regret of gradient descent, AdaGrad (Duchi et al., 2011), mirror descent and a variety of related algorithms satisfy (1), albeit with different constant terms that are hidden in the big-$O$ notation. Finally, the $1/\sqrt{N}$ rate is also a *lower-bound*. It cannot be improved without additional assumptions.

### 2.3. Neural Taylor Approximation

Consider a network with $L-1$ hidden layers and weight matrices $\mathbf{W} := \{\mathbf{W}_1, \dots, \mathbf{W}_L\}$. Let $\mathbf{x}_0$ denote the input. For hidden layers $l \in \{1, \dots, L-1\}$, set $\mathbf{a}_l = \mathbf{W}_l \cdot \mathbf{x}_{l-1}$ and $\mathbf{x}_l = s(\mathbf{a}_l)$ where $s(\cdot)$ is applied coordinatewise. The last layer outputs $\mathbf{x}_L = \mathbf{a}_L = \mathbf{W}_L \cdot \mathbf{x}_{L-1}$. Let $p_l$ denote the size of the $l^{\text{th}}$ layer; $p_0$ is the size of the input and $p_L$ is the size of the output. Suppose the loss $\ell(\mathbf{f}, y)$ is smooth and convex in the first argument. The training data is $(\mathbf{x}^d, y^d)_{d=1}^D$. The network is trained on stochastic samples from the training data on a series of rounds $n = 1, \dots, N$. For simplicity we assume minibatch size 1; the results generalize without difficulty.

We recall backprop using notation from Martens et al. (2012). Let $\mathbf{J}_{\mathbf{b}}^{\mathbf{a}}$ denote the Jacobian matrix of the vector $\mathbf{a}$ with respect to the vector $\mathbf{b}$. By the chain rule the gradient decomposes as

$$\begin{aligned}\nabla_{\mathbf{W}_l} \ell(\mathbf{f}_{\mathbf{W}}(\mathbf{x}_0), y) &= \underbrace{\mathbf{J}_L^{\mathcal{E}} \cdot \mathbf{J}_{L-1}^L \cdots \mathbf{J}_l^{l+1}}_{\boldsymbol{\delta}_l} \otimes \mathbf{x}_{l-1} \\ &= \underbrace{\mathbf{J}_L^{\mathcal{E}}}_{\nabla_{\mathbf{f}} \ell(\mathbf{f},y)} \cdot \underbrace{\mathbf{J}_l^L \otimes \mathbf{x}_{l-1}}_{\nabla_{\mathbf{W}_l} \mathbf{f}_{\mathbf{W}}(\mathbf{x}_0) =: \mathbf{G}_l}\end{aligned} \tag{2}$$

where $\boldsymbol{\delta}_l = \mathbf{J}_l^{\mathcal{E}}$ is the backpropagated error computed recursively via $\boldsymbol{\delta}_l = \boldsymbol{\delta}_{l+1} \cdot \mathbf{J}_l^{l+1}$.[2] The middle expression in (2) is the standard representation of backpropagated gradients. The expression on the right factorizes the backpropagated error $\boldsymbol{\delta}_l = \mathbf{J}_L^{\mathcal{E}} \cdot \mathbf{J}_l^L$ into the gradient of the loss $\mathbf{J}_L^{\mathcal{E}}$ and the Jacobian $\mathbf{J}_l^L$ between layers, which describes gradient flow *within* the network.

The first-order Taylor approximation to a differentiable function $f : \mathbb{R} \to \mathbb{R}$ near $a$ is $T_a(x) = f(a) + f'(a) \cdot (x - a)$. The neural Taylor approximation for a fully connected network is as follows.

**Definition 1.** *The* ***Jacobian tensor of layer*** $l$, $\mathbf{G}_l := \mathbf{J}_l^L \otimes \mathbf{x}_{l-1}$, *is the gradient of the output of the neural network with respect to the weights of layer $l$. It is the outer product of a $(p_L \times p_l)$-matrix with a $p_{l-1}$-vector, and so is a $(p_L, p_l, p_{l-1})$-tensor.*

*Given $\mathbf{G}_l$ and $(p_l \times p_{l-1})$-matrix $\mathbf{V}$, the expression $\langle \mathbf{G}_l, \mathbf{V} \rangle := \mathbf{J}_l^L \cdot \mathbf{V} \cdot \mathbf{x}_{l-1}$ is the $p_L$-vector computed via matrix-matrix-vector multiplication. The* ***neural Taylor approximation*** *to* $\mathbf{f}$ *in a neighborhood of* $\mathbf{W}^n$*, given input* $\mathbf{x}_0^n$,

[2] Note: we suppress the dependence of the Jacobians on the round $n$ to simplify notation.

*is the first-order Taylor expansion*

$$\mathbf{T}^n(\mathbf{V}) := \mathbf{f}_{\mathbf{W}^n}(\mathbf{x}_0^n) + \sum_{l=1}^{L} \langle \mathbf{G}_l, \mathbf{V}_l - \mathbf{W}_l^n \rangle \approx \mathbf{f}_{\mathbf{V}}(\mathbf{x}_0^n).$$

*Finally, the* ***Taylor loss*** *of the network on round $n$ is* $\mathfrak{T}^n(\mathbf{V}) = \ell(\mathbf{T}^n(\mathbf{V}), y^n)$.

The **Taylor approximation to layer** $l$ is

$$\mathbf{T}_l^n(\mathbf{V}_l) := \mathbf{f}_{\mathbf{W}^n}(\mathbf{x}_0^n) + \langle \mathbf{G}_l, \mathbf{V}_l - \mathbf{W}_l^n \rangle.$$

We can also construct the Taylor approximation to neuron $\alpha$ in layer $l$. Let the $p_L$-vector $\mathbf{J}_\alpha^L := \mathbf{J}_l^L[:, \alpha]$ denote the Jacobian with respect to neuron $\alpha$ and let the $(p_L \times p_{l-1})$-matrix $\mathbf{G}_\alpha := \mathbf{J}_\alpha^L \otimes \mathbf{x}_{l-1}$ denote the Jacobian with respect to the weights of neuron $\alpha$. The **Taylor approximation to neuron** $\alpha$ is

$$\mathbf{T}_\alpha^n(\mathbf{V}_\alpha) := \mathbf{f}_{\mathbf{W}^n}(\mathbf{x}_0^n) + \langle \mathbf{G}_\alpha, \mathbf{V}_\alpha - \mathbf{W}_\alpha^n \rangle.$$

The Taylor losses are the simplest non-trivial (i.e. non-affine) convex functions encoding the information generated by backprop, see section A4.

The following theorem provides convergence guarantees at mutiple spatial scales: network-wise, layer-wise and neuronal. See sections A2 for a proof of the theorem. It is not currently clear which scale provides the tightest bound.

**Theorem 2** (no-regret relative to Taylor optimum)**.** *Suppose, as in Theorem 1, the Taylor losses have bounded gradients and the weights of the neural network have bounded diameter during training.*

*Suppose the neural net is optimized by an algorithm with* $\texttt{Regret}(N) \leq O(\sqrt{N})$ *such as gradient descent, AdaGrad, Adam or mirror descent.*

- **Network guarantee:** *The running average of the training error of the neural network satisfies*

$$\underbrace{\frac{1}{N}\sum_{n=1}^{N} \ell\left(f_{\mathbf{W}^n}(\mathbf{x}_0^n), y^n\right)}_{\text{running average of training errors}} \leq \underbrace{\min_{\mathbf{V}} \left\{ \frac{1}{N}\sum_{n=1}^{N} \mathfrak{T}^n(\mathbf{V}) \right\}}_{\text{Taylor optimum}} + \underbrace{O\left(\frac{1}{\sqrt{N}}\right)}_{\texttt{Regret}(N)/N}. \quad (3)$$

- **Layer-wise / Neuron-wise guarantee:** *The Taylor loss of [layer-$l$ / neuron-$\alpha$] is* $\mathfrak{T}_{l/\alpha}^n(\mathbf{V}_{l/\alpha}) := \ell(\mathbf{T}_{l/\alpha}^n(\mathbf{V}_{l/\alpha}), y^n)$*. Then,*

$$\underbrace{\frac{1}{N}\sum_{n=1}^{N} \ell\left(f_{\mathbf{W}^n}(\mathbf{x}_0^n), y^n\right)}_{\text{running average of training errors}} \leq \underbrace{\min_{\mathbf{V}_{l/\alpha}} \left\{ \frac{1}{N}\sum_{n=1}^{N} \mathfrak{T}_{l/\alpha}^n(\mathbf{V}_{l/\alpha}) \right\}}_{\text{layer-wise/neuronal Taylor optimum}} \quad (4)$$
$$+ \underbrace{O\left(\frac{1}{\sqrt{N}}\right)}_{\texttt{Regret}(N)/N}.$$

The running average of errors during training (or cumulative loss) is the same quantity that arises in the analyses of Adam and Adagrad (Kingma & Ba, 2015; Duchi et al., 2011).

**Implications of the theorem.** The global optima of neural nets are not computationally accessible. Theorem 2 sidesteps the problem by providing a guarantee relative to the Taylor optimum. The bound is *path-dependent*; it depends on the convex snapshots encountered by backprop during training.

Path-dependency is a key feature of the theorem. It is a simple matter to construct a deep fully connected network ($>$ 100 layers) that fails to learn because gradients do not propagate through the network (He et al., 2016). A convergence theorem for neural nets must also be applicable in such pathological cases. Theorem 2 still holds because the failure of gradients to propagate through the network results in Taylor losses with poor solutions.

Although the bound in theorem 2 is path-dependent, it is nevertheless meaningful. The right-hand-side is given by the Taylor optimum, which is the optimal solution to the best convex approximations to the actual losses; best in the sense that they have the same value and have the same gradient for the encountered weights. The theorem replaces a seemingly intractable problem – neither smooth nor convex – with a sequence of convex problems.

Empirically, see below, we find that the Taylor optimum is a tough target on a range of datasets and settings: MNIST and CIFAR10; supervised and unsupervised learning; convolutional and fully-connected architectures; under a variety of optimizers (Adam, SGD, RMSProp), and for individual neurons as well as entire layers.

Finally, the decomposition of learning over rectifier networks into vanilla optimization and exploration components suggests investigating the exploratory behavior of different optimizers – with the theorem providing concrete tools to do so, see section 3.

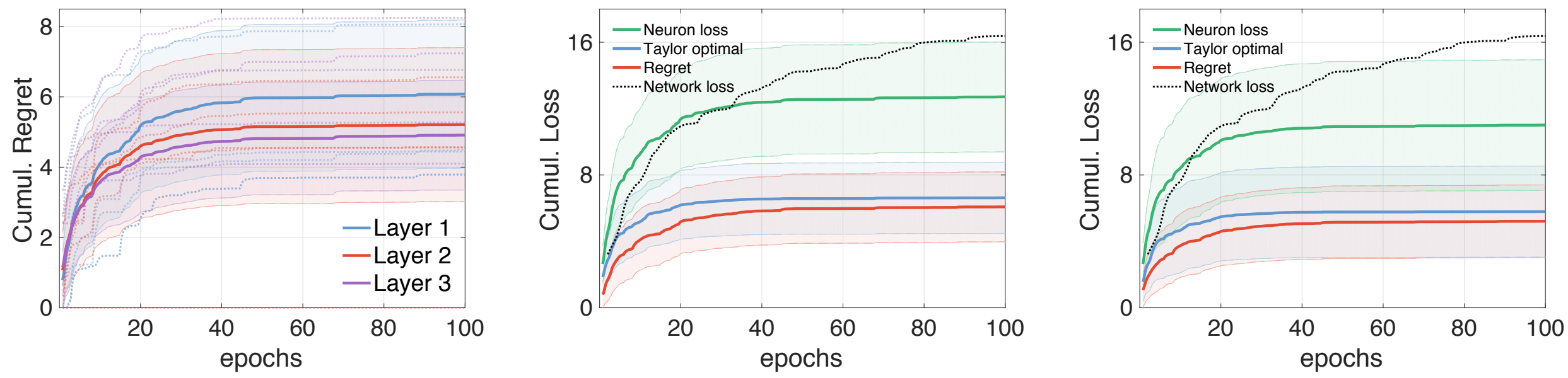

(a) All conv layers (b) Layer 1 (input layer) (c) Layer 2

Fig. 3: **Average normalized cumulative regret for RMSProp on CIFAR-10.** (a) Average regret incurred by neurons in each layer over 50 neurons/layer. (b)-(c) Average regret incurred eachs neuron in layers 1 and 2 respectively, along with average loss, Taylor optimum and cumulative network loss. Shaded areas represent one standard deviation.

### 2.4. Empirical Analysis of Online Neural Optimization

This section empirically investigates the Taylor optimum and regret terms in theorem 2 on two tasks:

**Autoencoder trained on MNIST.** Dense layers with architecture $784 \to 50 \to 30 \to 20 \to 30 \to 50 \to 784$ and ReLU non-linearities. Trained with MSE loss using minibatches of 64.

**Convnet trained on CIFAR-10.** Three convolutional layers with stack size 64 and $5 \times 5$ receptive fields, ReLU non-linearities and $2 \times 2$ max-pooling. Followed by a 192 unit fully-connected layer with ReLU before a ten-dimensional fully-connected output layer. Trained with cross-entropy loss using minibatches of 128.

For both tasks we compare the optimization performance of Adam, RMSProp and SGD (figure 6 in appendix). Learning rates were tuned for optimal performance. Additional parameters for Adam and RMSProp were left at default. For the convnet all three methods perform equally well: achieving a small loss and an accuracy of $\geq 99\%$ on the training set. However, SGD exhibits slightly more variance. For the autoencoder, although it is an extremely simple model, SGD with the best (fixed) learning rate performs significantly worse than the adaptive optimizers.

The neuronal and layer-wise regret are evaluated for each model. At every iteration we record the training error – the left-hand-side of eq. (4). To evaluate the Taylor loss, we record the input to the neuron/layer, its weights, the output of the network and the gradient tensor $\mathbf{G}_l$. After training, we minimize the Taylor loss with respect to $\mathbf{V}$ to find the Taylor optimum at each round. The regret is the difference between the observed training loss and the optimal Taylor loss.

The figures show *cumulative* losses and regret. For illustrative purposes we normalize by $1/\sqrt{N}$: quantities growing at $\sqrt{N}$ therefore flatten out. Figure 3(a) compares the average regret incurred by *neurons* in each convolutional layer of the convnet. Shaded regions show one standard deviation. Dashed lines are the regret of individual neurons – importantly the regret behaviour of neurons holds both on average and individually. Figs 3(b) and 3(c) show the regret, cumulative loss incurred by the network, the average loss incurred and the Taylor optimal loss for neurons in layers 1 and 2 respectively.

Fig. 4 compares Adam, RMSProp and SGD. Figure 4(a) shows the *layer-wise* regret on the convnet. The regret of all of the optimizers scales as $\sqrt{N}$ for both models, matching the bound in Theorem 2. The additional variance exhibited by SGD explains the difference in regret magnitude. Similar behaviour was observed in the other layers of the networks and also for convnets trained on MNIST.

Figure 4(b) shows the same plot for the autoencoder. The regret of all methods scales as $\sqrt{N}$ (this also holds for the other layers in the network). The gap in performance can be further explained by examining the difference between the observed loss and Taylor optimal loss. Figure 4(c) compares these quantities for each method on the autoencoder. The adaptive optimizers incur lower losses than SGD. Further, the gap between the actually incurred and optimal loss is smaller for adaptive optimizers. This is possibly because adaptive optimizers find better *activation configurations* of the network, see discussion in section 3.

Remarkably, figures 3 and 4 confirm that regret scales as $\sqrt{N}$ for a variety of optimizers, datasets, models, neurons and layers – verifying the multi-scale guarantee of Theorem 2. A possible explanation for why optimizers match the worst-case $(1/\sqrt{N})$ regret is that the adversary (that is, the landscape) keeps revealing Taylor losses with better solutions. The optimizer struggles to keep up with the hindsight optimum on the constantly changing Taylor losses.

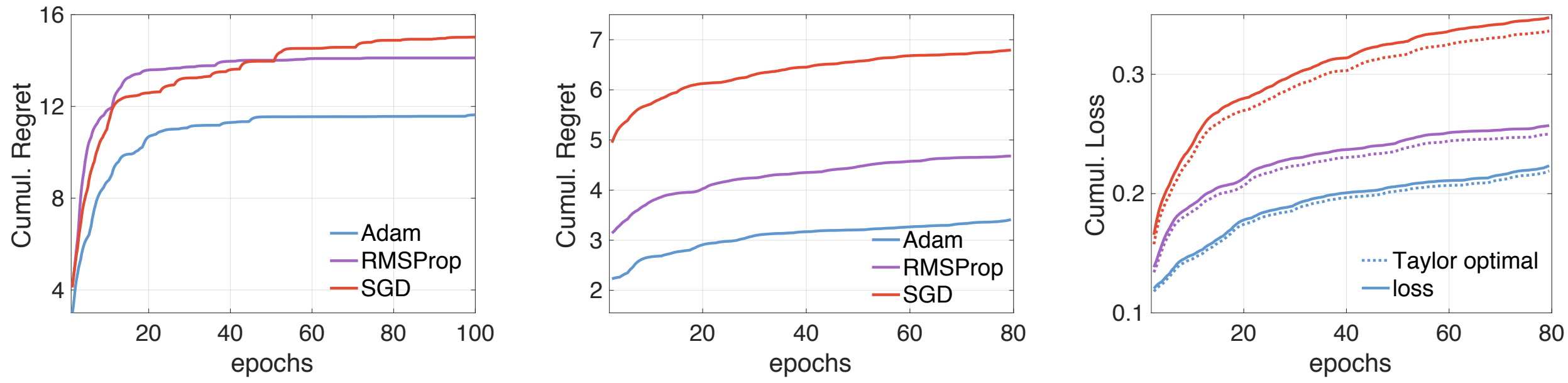

(a) Convnet regret, layer 2 (b) Autoencoder regret, layer 3 (c) Cumul. losses for autoencoder

Fig. 4: **Comparison of regret for Adam, RMSProp and SGD.** The $y$-axis in (b) is scaled by $\times 1000$. (c) reports cumulative loss and Taylor optimal loss on layer 3 for each method.

## 3. Optimization and Exploration in Rectifier Neural Networks

Poor optima in rectifier nets are related to shattered gradients: backprop cannot estimate gradients in nearby smooth regions without directly computing them; the flashlight does not shine across kinks. Two recent papers have shown that noise improves the local optima found during training: Neelakantan et al. (2016) introduce noise into gradients whereas Gulcehre et al. (2016) use noisy activations to extract gradient information from across kinks. Intuitively, noise is a mechanism to "peer around kinks" in shattered landscapes.

Introducing noise is not the only way to find better optima. Not only do adaptive optimizers often converge *faster* than vanilla gradient descent, they often also converge to *better* local minima.

This section investigates how adaptive optimizers explore shattered landscapes. Section 3.1 shows that smooth regions in rectifier nets correspond to configurations of active neurons and that neural Taylor approximations clamp the activation configuration – i.e. the convex flashlight does not shine across kinks in the landscape. Section 3.2 observes that adaptive optimizers incorporate an exploration bias and hypothesizes that the success of adaptive optimizers derives from exploring the set of smooth regions more extensively than SGD. Section 3.3 evaluates the hypothesis empirically.

### 3.1. The Role of Activation Configurations in Optimization

We describe how configurations of active neurons relate to smooth regions of rectifier networks and to neural Taylor approximations. Recall that the loss of a neural net on its training data is

$$\hat{\ell}(\mathbf{W}) = \frac{1}{D}\sum_{d=1}^{D} \ell(f_{\mathbf{W}}(\mathbf{x}^d), y^d).$$

**Definition 2.** *Enumerate the data as $[D] = \{1, \ldots, D\}$ and neurons as $[M]$. The* ***activation configuration*** *$\boldsymbol{\mathcal{A}}(\mathbf{W})$ is a $(D \times M)$ binary matrix representing the active neurons for each input. The set of all possible activation configurations corresponds to the set of all $(D \times M)$ binary matrices.*

**Observation 1** (activation configurations correspond to smooth regions in rectifier networks)**.**
*A parameter value exhibits a kink in $\hat{\ell}$ iff an infinitesimal change alters the of activation configuration, i.e. $\hat{\ell}$ is not smooth at $\mathbf{W}$ iff there is a $\mathbf{V}$ s.t. $\boldsymbol{\mathcal{A}}(\mathbf{W}) \neq \boldsymbol{\mathcal{A}}(\mathbf{W} + \delta\mathbf{V})$ for all $\delta > 0$.*

The neural Taylor approximation to a rectifier net admits a natural description in terms of activation configurations.

**Observation 2** (the Taylor approximation clamps activation configurations in rectifier networks)**.**
*Suppose datapoint $d$ is sampled on round $n$. Let $\mathbf{1}_k := \boldsymbol{\mathcal{A}}(\mathbf{W}^n)[d, \texttt{layer } k]$ be the $p_k$-vector given by entries of row $d$ of $\boldsymbol{\mathcal{A}}(\mathbf{W}^n)$ corresponding to neurons in layer $k$ of a rectifier network. The Taylor approximation $\mathbf{T}^n_l$ is*

$$\mathbf{T}^n_l(\mathbf{V}_l) = \underbrace{\mathbf{1}_l \cdot \left(\prod_{k\neq l} \mathbf{W}^n_k \cdot \text{diag}(\mathbf{1}_k)\right)}_{\textit{clamped}} \cdot \underbrace{(\mathbf{V}_l - \mathbf{W}^n_l)}_{\textit{free}}$$

*which clamps the activation configuration, and weights of all layers excluding $l$.*

Observations 1 and 2 connect shattered gradients in rectifier nets to activation configurations and the Taylor loss. The main implication is to factorize neural optimization into hard (finding "good" smooth regions) and easy (optimizing within a smooth region) subproblems that correspond, roughly, to finding "good" Taylor losses and optimizing them respectively.

### 3.2. RMS-Normalization encourages Exploration

Adaptive optimizers based on root-mean-square normalization exhibit an up-to-exponential improvement over non-adaptive methods when gradients are sparse (Duchi et al., 2013) or low-rank (Krummenacher et al., 2016) in convex settings. We propose an alternate explanation for the performance of adaptive optimizers in nonconvex nonsmooth settings.

Let $\overline{\nabla}\ell := \frac{1}{D}\sum_{d=1}^{D} \nabla \ell_d$ denote the average gradient over a dataset. RProp replaces the average gradient with its coordinatewise sign (Riedmiller & Braun, 1993). An interesting characterization of the signed-gradient is

**Observation 3** (signed-gradient is a maximizer)**.** *Suppose none of the coordinates in $\overline{\nabla}\ell$ are zero. The signed-gradient satisfies*

$$\texttt{sign}(\overline{\nabla}\ell) = \operatorname*{argmax}_{\mathbf{x}\in B^p_\infty} \Big\{ \|\mathbf{x}\|_1 \,:\, \langle \mathbf{x}, \overline{\nabla}\ell\rangle \geq 0 \Big\},$$
$$\textit{where } B^p_\infty = \{\mathbf{x}\in\mathbb{R}^p : \max_{i=1,\ldots,p} |x_i| \leq 1\}.$$

The signed-gradient therefore has two key properties. Firstly, small weight updates using the signed-gradient decrease the loss since $\langle \overline{\nabla}\ell, \texttt{sign}(\overline{\nabla}\ell)\rangle > 0$. Secondly, the signed-gradient is the update that, subject to an $\ell_\infty$ constraint, has the largest impact on the most coordinates. To adapt RProp to minibatches, Hinton and Tieleman suggested to approximate the signed gradient by normalizing with the root-mean-square: $\texttt{sign}(\overline{\nabla}\ell) \approx \frac{\sum_{d=1}^{D}\nabla \ell_d}{\sqrt{\sum_{d=1}^{D}(\nabla \ell_d)^2}}$, where $(\nabla \ell_d)^2$ is the square taken coordinatewise. Viewing the signed-gradient as changing weights – or exploring – maximally suggests the following hypothesis:

**Hypothesis 1** (RMS-normalization encourages exploration over activation configurations)**.** *Gradient descent with RMS-normalized updates (or running average of RMS) performs a broader search through the space of activation configurations than vanilla gradient descent.*

### 3.3. Empirical Analysis of Exploration by Adaptive Optimizers

Motivated by hypothesis 1, we investigate how RMSProp and SGD explore the space of activation configurations on the tasks from section 2.4; see A6 for details.

For fixed parameters $\mathbf{W}$, the activation configuration of a neural net with $M$ neurons and $D$ datapoints is represented as a $(D \times M)$ binary matrix, recall definition 2. The set of activation configurations encountered by a network over $N$ rounds of training is represented by an $(N, D, M)$ binary tensor denoted $\boldsymbol{\mathcal{A}}$ where $\boldsymbol{\mathcal{A}}_n := \boldsymbol{\mathcal{A}}[n, :, :] := \boldsymbol{\mathcal{A}}(\mathbf{W}^n)$.

Figure 5 quantifies exploration in the space of activation configurations in three ways:

*5(a): Hamming distance* plots $\min_{k<n} \|\boldsymbol{\mathcal{A}}_n - \boldsymbol{\mathcal{A}}_k\|_F^2$, the minimum Hamming distance between the current activation configuration and all previous configurations. It indicates the novelty of the current activation configuration.

*5(b): Activation-state switches* plots $\frac{1}{\texttt{tot}}\sum_{n=1}^{N-1}\big\|\boldsymbol{\mathcal{A}}_n[d, :] - \boldsymbol{\mathcal{A}}_{n-1}[d, :]\big\|_F^2$, the total number of times each data point (sorted) switches its activation state across all neurons and epochs as a proportion of possible switches. It indicates the variability of the network response.

*5(c): Log-product of singular values.* The matrix $\boldsymbol{\mathcal{A}}[:, :, m]$ specifies the rounds and datapoints that activate neuron $m$. The right column plots the log-product of $\boldsymbol{\mathcal{A}}[:, :, m]$'s first 50 singular values for each neuron (sorted).[3] It indicates the (log-)volume of configuration space covered by each neuron. Note that values reaching the bottom of the plot indicate singular values near 0.

We observe the following.

**RMSProp explores the space of activation configurations far more than SGD.** The result holds on both tasks, across all three measures, and for multiple learning rates for SGD (including the optimally tuned rate). The finding provides evidence for hypothesis 1.

**RMSProp converges to a significantly better local optimum on the autoencoder**, see Fig. 6. We observe no difference on CIFAR-10. We hypothesize that RMSProp finds a better optimum through more extensive exploration through the space of activation configurations. CIFAR is an easier problem and possibly requires less exploration.

**Adam explores less than RMSProp.** Adam achieves the best performance on the autoencoder. Surprisingly, it explores substantially less than RMSProp according to the Hamming distance and activation-switches, although still more than SGD. The singular values provide a higher-resolution analysis: the $\pm 40$ most exploratory neurons match the behavior of RMSProp, with a sharp dropoff from neuron 60 onwards. A possible explanation is that momentum encourages *targeted* exploration by rapidly discarding avenues that are not promising. The results for Adam are more ambiguous than for RMSProp compared to SGD.

[3] The time-average is subtracted from each column of $\boldsymbol{\mathcal{A}}[:, :, m]$. If the response of neuron $m$ to datapoint $d$ is constant over all rounds, then column $\boldsymbol{\mathcal{A}}[:, d, m]$ maps to $(0, \ldots, 0)$ and does not contribute to the volume.

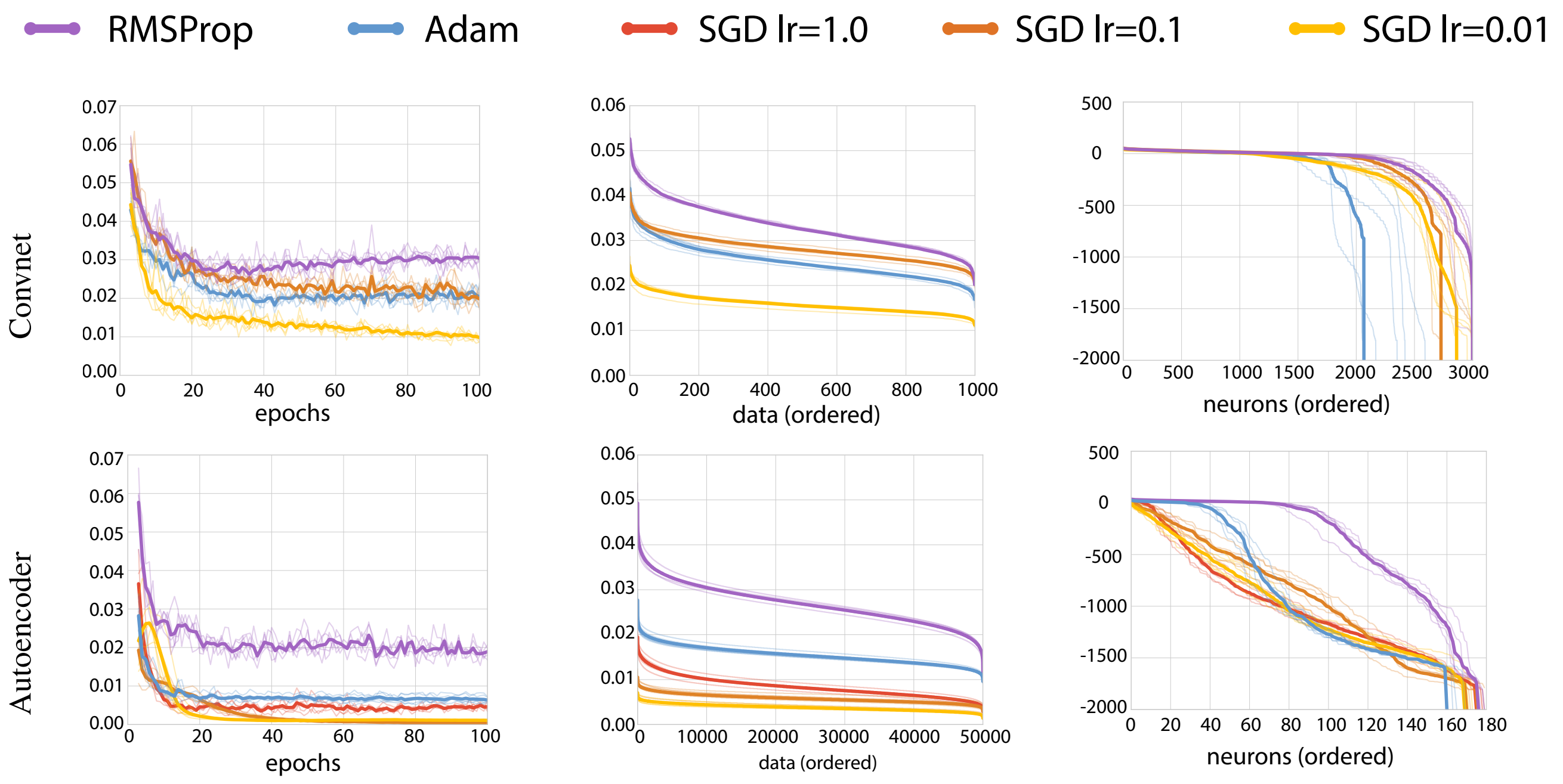

(a) Minimum hamming distance. (b) Activation-state switches. (c) Log-product of singular values.

Fig. 5: **Top: results for a CIFAR-trained convnet. Bottom: MNIST-trained autoencoder.** (a) Minimum hamming distance between the activation configuration at curent epoch and all previous epochs. (b) Number of activation-state switches undergone for all neurons over all epochs for each data point (sorted). (c) Log-product of the first 50 singular values of each neuron activation configuration (sorted).

More generally, the role of momentum in nonsmooth nonconvex optimization requires more investigation.

## 4. Discussion

Rectifier convnets are the dominant technology in computer vision and a host of related applications. Our main contribution is the first convergence result applicable to convnets as they are used in practice, including rectifier nets, max-pooling, dropout and related methods. The key analytical tool is the neural Taylor approximation, the first-order approximation to the output of a neural net. The Taylor loss – the loss on the neural Taylor approximation – is a convex approximation to the loss of the network. Remarkably, the convergence rate matches known lower bounds on *convex* nonsmooth functions (Bubeck, 2015). Experiments in section 2.4 show the regret matches the theoretical analysis under a wide range of settings.

The bound in theorem 2 contains an easy term to optimize (the regret) and a hard term (finding "good" Taylor losses). Section 3.1 observes that the Taylor losses speak directly to the fundamental difficulty of optimizing nonsmooth functions: that gradients are shattered – the gradient at a point is not a reliable estimate of nearby gradients.

Smooth regions of rectifier nets correspond to activation configurations. Gradients in one smooth region cannot be used to estimate gradients in another. Exploring the set of activation configurations may therefore be crucial for optimizers to find better local minima in shattered landscapes. Empirical results in section 3.3 suggest that the improved performance of RMSProp over SGD can be explained in part by a carefully tuned exploration bias.

Finally, the paper raises several questions:

1. To what extent is exploration necessary for good performance?
2. Can exploration/exploitation tradeoffs in nonsmooth neural nets be quantified?
3. There are exponentially more kinks in early layers (near the input) compared to later layers. Should optimizers explore more aggressively in early layers?
4. Can exploring activation configurations help design better optimizers?

The Taylor decomposition provides a useful tool for separating the convex and nonconvex aspects of neural optimization, and may also prove useful when tackling exploration in neural nets.

## Acknowledgements

We thank L. Helminger and T. Vogels for useful discussions and help with TensorFlow. Some experiments were performed using a Tesla K80 kindly donated by Nvidia.

# APPENDIX

## A1. Background on convex optimization

A continuous function $f$ is smooth if there exists a $\beta > 0$ such that $\| \nabla f(\mathbf{x}) - \nabla f(\mathbf{y}) \|_2 \leq \beta \cdot \|\mathbf{x} - \mathbf{y}\|_2$ for all $\mathbf{x}$ and $\mathbf{y}$ in the domain. Rectifiers are not smooth for any value of $\beta$.

**Nonsmooth convex functions.** Let $X \subset \mathbb{R}^p$ be a convex set contained in a ball of radius $R$. Let $\ell : X \to \mathbb{R}$ be a convex function. Section 3.1 of Bubeck (2015) shows that projected gradient descent has convergence guarantee

$$\ell\left(\frac{1}{N}\sum_{n=1}^{N} \mathbf{w}^n\right) - \ell(\mathbf{w}^*) \leq O\left(\frac{1}{\sqrt{N}}\right)$$

where $\mathbf{w}^n$ are generated by gradient descent and $\mathbf{w}^* := \text{argmin}_{\mathbf{w}\in X}\, \ell(\mathbf{w})$ is the minimizer of $\ell$. It is also shown, section 3.5, that

$$\min_{1\leq n\leq N} \ell(\mathbf{w}^n) - \ell(\mathbf{w}^*) \geq \Omega\left(\frac{1}{\sqrt{N}}\right)$$

where the weights are in the span of the previously observed gradients: $\mathbf{w}^n \in \text{span}\{\nabla \ell(\mathbf{w}^1), \ldots, \nabla \ell(\mathbf{w}^{n-1})\}$ for all $n \in \{1, \ldots, N\}$.

The gradient of a convex function increases monotonically. That is

$$\big\langle \nabla \ell(\mathbf{w}) - \nabla \ell(\mathbf{v}), \mathbf{w} - \mathbf{v} \big\rangle \geq 0$$

for all points $\mathbf{w}, \mathbf{v}$ where the gradient exists. Gradients at one point of a nonsmooth convex function therefore do contain information about other points, although not as much information as in the smooth case. In contrast, the gradients of nonsmooth nonconvex functions can vary wildly as shown in Fig. 1.

**Smooth convex functions.** In the smooth setting, gradient descent converges at rate $\frac{1}{N}$. The lower bound for convergence is even better, $\frac{1}{N^2}$. The lower bound is achieved by Nesterov's accelerated gradient descent method.

## A2. Proof of Theorem 2

*Proof.* We prove the network case; the others are similar. The Taylor loss has three key properties by construction:

T1. The Taylor loss $\mathfrak{T}^n$ coincides with the loss at $\mathbf{W}^n$:
$\ell(\mathbf{f}_{\mathbf{W}^n}(\mathbf{x}_0^n), y^n) = \mathfrak{T}^n(\mathbf{V})_{|\mathbf{V}=\mathbf{W}^n}$

T2. The Taylor loss gradient $\mathfrak{T}^n$ coincides with the loss gradient at $\mathbf{W}^n$:
$\nabla_{\mathbf{W}}\, \ell(\mathbf{f}_{\mathbf{W}^n}(\mathbf{x}_0^n), y^n) = \nabla_{\mathbf{V}}\, \mathfrak{T}^n(\mathbf{V})_{|\mathbf{V}=\mathbf{W}^n}$

T3. The Taylor losses are convex functions of $\mathbf{V}$ because $\ell(\mathbf{f}, y)$ is convex in its first argument and convexity is invariant under affine maps. If $\ell$ is a convex function, then so is $g(x) = \ell(Ax + b)$, where $A \in \mathbb{R}^{m\times n}$ and $b \in \mathbb{R}^m$.

By T1, the training loss, i.e. the left-hand side of (3), exactly coincides with the Taylor losses. By T2, the gradients of the Taylor losses exactly coincide with the errors computed by backpropagation on the training losses. That is, the training loss over $n$ rounds is indistinguishable from the Taylor losses to the first order:

$$\begin{aligned}
\text{losses: } & \frac{1}{N}\sum_{n=1}^{N} \ell\left(f_{\mathbf{W}^n}(\mathbf{x}_0^n), y^n\right) \\
& = \frac{1}{N}\sum_{n=1}^{N} \mathfrak{T}^n\left(\mathbf{W}^n\right) \\
\text{gradients: } & \nabla_{\mathbf{W}}\left(\frac{1}{N}\sum_{n=1}^{N} \ell\left(f_{\mathbf{W}^n}(\mathbf{x}_0^n), y^n\right)\right) \\
& = \nabla_{\mathbf{W}}\left(\frac{1}{N}\sum_{n=1}^{N} \mathfrak{T}^n\left(\mathbf{W}^n\right)\right)
\end{aligned}$$

We can therefore substitute the Taylor losses in place of the training loss $(f_{\mathbf{W}}(\mathbf{x}_0), y)$ without altering either the losses incurred during training or the dynamics of backpropagation (or any first-order method).

Since the Taylor losses are convex, the bound holds for any no-regret optimizer following Zinkevich (2003). □

## A3. Proof of Observations in Section 3

**Proof of observation 1.**

*Proof.* The loss $\ell(\mathbf{f}, y)$ is a smooth function of the network's output $\mathbf{f}$ by assumption. Kinks in $\hat{\ell}(\mathbf{W}) = \frac{1}{D}\sum_{d=1}^{D} \ell\big(f_{\mathbf{W}}(\mathbf{x}^d), y^d\big)$ can therefore only arise when a rectifier changes its activation for at least one element of the training data. □

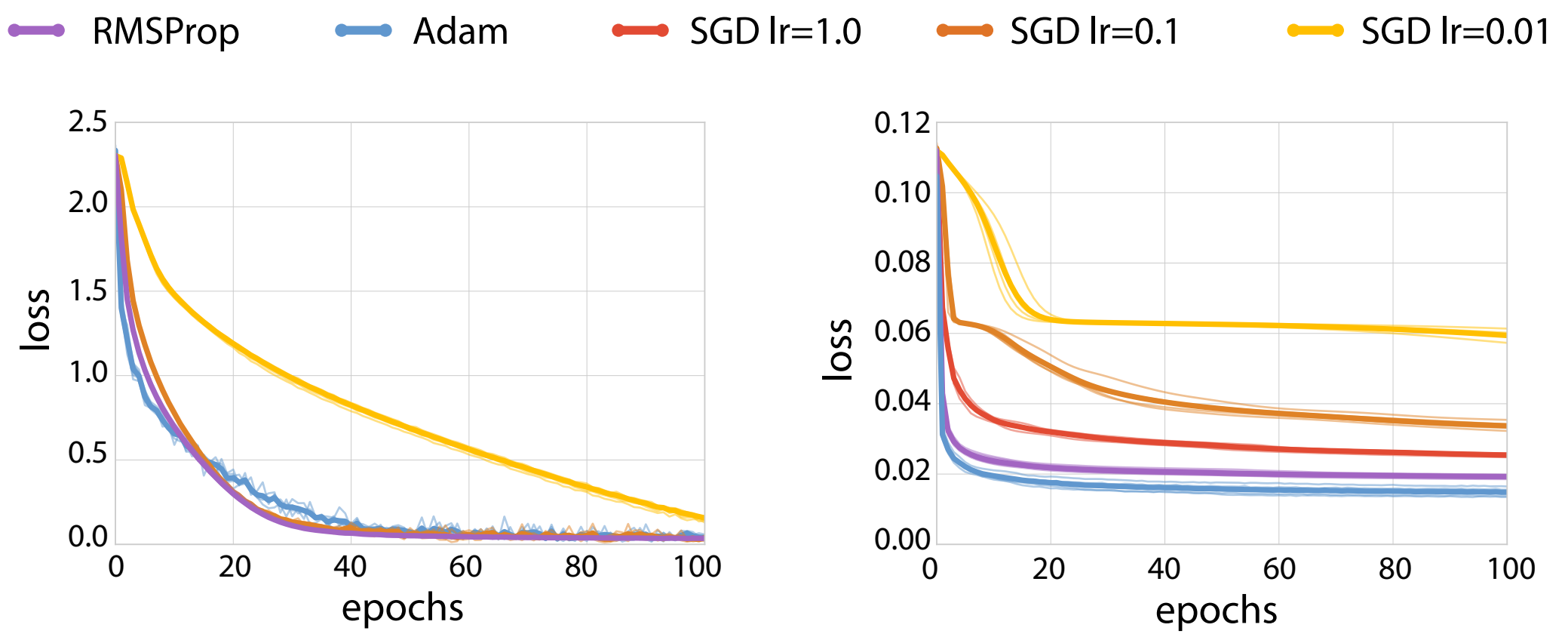

(b) Convnet

(c) Autoencoder

Fig. 6: **Training loss on CIFAR-10 and MNIST.**

**Proof of observation 2.** Note that the rectifier is $\rho(a) = \max(0, a)$ with derivative $\rho'(a) = 1$ if $a > 0$ and $\rho'(a) = 0$ if $a < 0$.

*Proof.* Recall that the Taylor approximation to layer $l$ is

$$\begin{aligned}\mathbf{T}_l^n(\mathbf{V}_l) &:= \mathbf{f}_{\mathbf{W}^n}(\mathbf{x}_0^n) + \langle \mathbf{G}_l, \mathbf{V}_l - \mathbf{W}_l^n \rangle \\ &= \mathbf{f}_{\mathbf{W}^n}(\mathbf{x}_0^n) + \mathbf{J}_l^L \cdot (\mathbf{V}_l - \mathbf{W}_l^n) \cdot \mathbf{x}_{l-1}^n \\ &= \mathbf{f}_{\mathbf{W}^n}(\mathbf{x}_0^n) + \left( \prod_{k=L}^{l+1} \mathbf{J}_{k-1}^k \right) \cdot (\mathbf{V}_l - \mathbf{W}_l^n) \cdot \mathbf{x}_{l-1}^n\end{aligned}$$

The Jacobian of layer $k$ is the function $\mathbf{J}_k^{k+1}(\mathbf{a}_k) = \mathbf{W}_{k+1} \cdot \text{diag}\left(s'(\mathbf{a}_k)\right)$ which in general varies nonlinearly with $\mathbf{a}_k$. The Taylor approximation *clamps* the Jacobian by setting it as constant.

For a layer of rectifiers, $s(\cdot) = \rho(\cdot)$, the Jacobian $\mathbf{J}_k^{k+1} = \mathbf{W}_{k+1} \cdot \text{diag}\left(\rho'(\mathbf{a}_k)\right)$ is constructed by zeroing out the rows of $\mathbf{W}_{k+1}$ corresponding to inactive neurons in layer $k$. It follows that the Taylor loss can be written as

$$\mathbf{T}_l^n(\mathbf{V}_l) = \left( \prod_{k=L}^{l+1} \mathbf{W}_k^n \cdot \text{diag}(\mathbf{1}_{k-1}) \right) \cdot (\mathbf{V}_l - \mathbf{W}_l^n) \cdot \mathbf{x}_{l-1}^n$$

Finally, observe that

$$\mathbf{x}_{l-1}^n = \left( \prod_{k=l-1}^{1} \text{diag}(\mathbf{1}_k) \cdot \mathbf{W}_k^n \right) \cdot \mathbf{x}_0^n$$

since $\text{diag}(\mathbf{1}_k) \cdot \mathbf{W}_k^n \cdot \mathbf{x}_{k-1}^n = \text{diag}\left(\rho'(\mathbf{a}_k)\right) \cdot \mathbf{W}_k^n \mathbf{x}_{k-1}^n = \rho(\mathbf{W}_k^n \cdot \mathbf{x}_{k-1}^n)$. □

**Proof of observation 3.**

*Proof.* Immediate. □

## A4. Comparison of Taylor loss with Taylor approximation to loss

It is instructive to compare the Taylor loss in definition 1 with the Taylor approximation to the loss. The Taylor loss is

$$\ell\Big(\mathbf{T}^n(\mathbf{V}), y^n\Big) = \ell\left(\mathbf{f}_{\mathbf{W}^n}(\mathbf{x}_0^n) + \sum_{l=1}^{L} \langle \mathbf{G}_l, \mathbf{V}_l - \mathbf{W}_l^n \rangle, y^n\right)$$

In contrast, the Taylor approximation to the loss is

$$\begin{aligned}T_\ell^n(\mathbf{V}) &= \underbrace{\ell\big(f_{\mathbf{W}^n}(\mathbf{x}_0^n), y^n\big)}_{\text{loss incurred on round } n} + \sum_{l=1}^{L} \big\langle \underbrace{\mathbf{J}_L^{\mathcal{E}} \cdot \mathbf{G}_l}_{\boldsymbol{\delta}_l}, \mathbf{V}_l - \mathbf{W}_l^n \big\rangle \\ &= \ell^n + \sum_{l=1}^{L} \big\langle \nabla_{\mathbf{W}_l} \ell, \mathbf{V}_l - \mathbf{W}_l^n \big\rangle.\end{aligned}$$

The constant term is the loss incurred on round $n$; the linear coefficients are the backpropagated errors.

It is easy to see that the two expressions have the same gradient. Why not work directly with the Taylor approximation to the loss? The problem is that the Taylor approximation to the loss is affine, and so decreases without bound. Upgrading to a second order Taylor approximation is no help since it is not convex.

## A5. Details on experiments on regret

See section 2.4 for the architecture of the autoencoder and convnet used. The hyperparameters used for different optimizers are as follows: the autoencoder uses learning rate $\eta = 0.001$ for RMSprop and $\eta = 0.01$ for Adam, while the convnet uses learning rate $\eta = 0.0005$ for RMSprop and $\eta = 0.0002$ for Adam. All other hyperparameters are kept at their literature-standard values.

Fig. 6 shows the training losses obtained by the convnet on CIFAR-10 and the autoencoder on MNIST.

The gradient tensor $\mathbf{G}_l$ is not computed explicitly by TensorFlow. Instead, it is necessary to compute the gradient of each component of the output layer (e.g. 10 in total for a network trained on CIFAR-10, 784 for an autoencoder trained on MNIST) with respect to $\mathbf{W}_l$ and then assemble the gradients into a tensor. When the loss is the squared error, the Taylor optimal at round $n$ can be computed in closed form. Otherwise we use SGD.

## A6. Details on experiments on exploration

Given matrix or vector $\mathbf{A}$ or $\mathbf{a}$, the squared Frobenius norm is

$$\|\mathbf{A}\|_F^2 = \sum_{m,n=1}^{M,N} A_{m,n}^2 \quad \text{and} \quad \|\mathbf{a}\|_F^2 = \sum_{n=1}^{N} a_n^2.$$

The Hamming distance between two binary vectors $\mathbf{a}$ and $\mathbf{b}$ can be computed as $\|\mathbf{a} - \mathbf{b}\|_F^2$.

For tractability in the convnet, we only record activations for $1\%$ of the CIFAR dataset, and at most 10000 units of each convolutional layer. We record the full network state on all inputs for the autoencoder. The singular value plots in figure 5 are calculated only on the first 50 epochs.